\documentclass[conference]{IEEEtran}
\IEEEoverridecommandlockouts
\usepackage{cite}
\usepackage{amsmath,amssymb,amsfonts}
\usepackage{algorithm}
\usepackage{algorithmic}
\usepackage{graphicx}
\usepackage{textcomp}
\usepackage{xcolor}

\def\BibTeX{{\rm B\kern-.05em{\sc i\kern-.025em b}\kern-.08em
    T\kern-.1667em\lower.7ex\hbox{E}\kern-.125emX}}
\begin{document}

\title{Passive Batch Injection Training Technique: Boosting Network Performance by Injecting Mini-Batches from a different Data Distribution}

\author{\IEEEauthorblockN{Pravendra Singh}
\IEEEauthorblockA{{Indian Institute of Technology Kanpur}\\
\tt\small psingh@iitk.ac.in}
\and
\IEEEauthorblockN{Pratik Mazumder}
\IEEEauthorblockA{{Indian Institute of Technology Kanpur}\\
\tt\small pratikm@iitk.ac.in}
\and
\IEEEauthorblockN{Vinay P. Namboodiri}
\IEEEauthorblockA{{Indian Institute of Technology Kanpur}\\
\tt\small vinaypn@iitk.ac.in}
}

\maketitle

\begin{abstract}
This work presents a novel training technique for deep neural networks that makes use of additional data from a  distribution that is different from that of the original input data.  This technique aims to reduce overfitting and improve the generalization performance of the network. Our proposed technique, namely Passive Batch Injection Training Technique (PBITT), even reduces the level of overfitting in networks that already use the standard techniques for reducing overfitting such as $L_2$ regularization and batch  normalization, resulting in significant accuracy improvements. Passive Batch Injection Training Technique (PBITT) introduces a few \textbf{passive} mini-batches into the training process that contain data from a distribution that is different from the input data distribution. This technique does not increase the number of parameters in the final model and also does not increase the inference (test) time but still improves the performance of deep CNNs. To the best of our knowledge, this is the first work that makes use of different data distribution to aid the training of convolutional neural networks (CNNs). We thoroughly evaluate the proposed approach on standard architectures: VGG, ResNet, and WideResNet, and on several popular datasets: CIFAR-10, CIFAR-100, SVHN, and ImageNet. We observe consistent accuracy improvement by using the proposed technique. We also show experimentally that the model trained by our technique generalizes well to other tasks such as object detection on the MS-COCO dataset using Faster R-CNN. We present extensive ablations to validate the proposed approach. Our approach improves the accuracy of VGG-16 by a significant margin of 2.1\% over the CIFAR-100 dataset.
\end{abstract}


\section{Introduction}
Deep neural networks have been immensely successful in many tasks. However, training them still remains very tricky. Several reasons ranging from vanishing/exploding gradients \cite{glorot2010understanding}, to feature statistic shifts \cite{ioffe2015batch}, to the proliferation of saddle points \cite{dauphin2014identifying}, to overfitting \cite{tetko1995neural,srivastava2014dropout}, and others, are responsible for this matter. Researchers have proposed several solutions to deal with these issues, examples of which include parameter initialization \cite{saxe2013exact}, residual connections \cite{he2016deep}, normalization of internal activations \cite{ioffe2015batch}, second-order optimization algorithms \cite{dauphin2014identifying}, and regularization techniques \cite{srivastava2014dropout,bansal2018can,ng2004feature,van2017l2}.

Various works \cite{singh2019hetconv,singh2019stability,singh2018leveraging,singh2019multi,singh2019accuracy,singh2019falf,singh2019hetconvijcv,mazumder2019cpwc,singh2019play} have been proposed for efficient deep learning. The work in \cite{bansal2018can} improves the performance of state-of-the-art models by using orthogonality regularization.  
While many training techniques use augmented data or additional data from the same source or produced synthetically, to improve the performance of networks, our proposed technique makes use of controlled injection of data mini-batches from a different data distribution. We refer to our proposed technique as the Passive Batch Injection Training Technique (PBITT). In this technique, during the training process, we introduce mini-batches of data (passive mini-batches) that are from a different distribution than the input data and also train on them. PBITT trains the network on both the passive and original mini-batches. 
This forces the model to search for a representation that captures the information needed for both types of mini-batches. 

However, since the passive mini-batch has a different data distribution, ideally, this setup should harm the performance of the network. But by controlling the ratio of passive mini-batches to original mini-batches, we are able to achieve a higher reduction in overfitting while negating the adverse effects of this setup and therefore achieve better performance. The passive mini-batch injection is controlled in such a way that the number of original mini-batches vastly outnumbers them. In practice, we draw the passive mini-batches from a dataset where the images do not match the general content or pattern of the original images at a specific size. This ensures that the distribution of passive mini-batches is different from the original input data.

The works presented in \cite{li2018learning,mallya2018packnet,mallya2018piggyback,he2018multi} add multiple tasks to a neural network with minimal loss in accuracy. However, their objective is to perform equally well in all these tasks using the same network. This, however, usually results in lower accuracy when compared to the accuracy of separate individual networks trained for one task each, if the tasks are from different data distributions. Our technique focuses on increasing the performance of the network on the original data only. The performance on the passive mini-batches will not be good since such mini-batches are an extreme minority when compared to the original mini-batches. 

The networks, we experiment on, already use batch normalization \cite{ioffe2015batch} and weight decay ($L_2$ regularization) to reduce overfitting, and we observe that our approach reduces the level of overfitting further, which results in accuracy improvement (refer to experimental section~\ref{experimental} and section~\ref{subsec:overfitting}). Our approach achieves these results without increasing the number of parameters or the inference (test) time of the final model.

As mentioned earlier, Passive Batch Injection Training Technique (PBITT) involves introducing a few passive mini-batches into the training process alongside the original mini-batches. We thoroughly analyze the effects of changing the number of original mini-batches for every passive mini-batch in our detailed ablation studies. We also analyze the effect of changing the datasets of our passive mini-batches in our ablation studies and experimentally show that our approach of Passive Batch Injection Training Technique (PBITT) does not depend on the selection of the passive dataset. We show experimentally that our approach improves over several recent benchmark results in classification on ImageNet, SVHN, and CIFAR datasets and generalizes well for object detection on the MS-COCO dataset.

The following are our contributions:
\begin{itemize}
    \item We propose a training technique: Passive Batch Injection Training Technique (PBITT), which significantly improves the performance of CNNs by using mini-batches of data from a distribution that is different from the original data distribution.
    \item Our approach improves the CNN performance without increasing the number of parameters or inference (test) time in the final model. 
    \item Our approach reduces overfitting and improves the generalization performance of networks, including those which already use the standard techniques for reducing overfitting. 
    \item We show that our proposed approach works well for various networks not only for classification but also for detection.
\end{itemize}

\section{Related Works}

Training of deep neural networks suffers from several tricky issues such as vanishing gradients, overfitting, unstable gradient, co-variate shift and others. Multiple strategies and techniques have been proposed to alleviate such issues such as parameter initialization \cite{saxe2013exact}, residual connections \cite{he2016deep}, normalization of internal activations \cite{ioffe2015batch}, second-order optimization algorithms \cite{dauphin2014identifying}, and regularization techniques \cite{srivastava2014dropout,bansal2018can,ng2004feature,van2017l2}.

\cite{glorot2010understanding,he2015delving} propose to enforce close to constant variances of each layer's output for initialization to reduce the problems of unstable gradient and covariate shift. \cite{ioffe2015batch} proposes to enforce identical distributions of each layer's output to reduce the internal covariate shift. \cite{salimans2016weight} proposes to decouple the norm of the weight vector from its direction to make the optimization easier. Orthogonal weights have been extensively studied in recurrent neural networks (RNNs) \cite{pascanu2013difficulty,dorobantu2016dizzyrnn,arjovsky2016unitary,mhammedi2017efficient,vorontsov2017orthogonality,wisdom2016full} to help avoid gradient vanishing/explosion. 

\cite{bansal2018can} improves CNN training and performance by using orthogonality regularizations during training deep CNNs and using tools like mutual coherence and restricted isometry property (SRIP). These orthogonality regularizations have a plug-and-play nature and can be easily incorporated into the training process of CNNs.

Performance improvement through the architectural improvement of CNNs has remained a hot topic from some time now \cite{huang2017densely,zoph2018learning}. The focus has been on design changes that improve the performance of networks on various tasks. Architectures like Inception models \cite{szegedy2015going} and VGGNet \cite{simonyan2014very} demonstrate how the quality of representation learned by a network can be significantly improved by increasing the depth of the network. While deeper architectures improved the performance of CNNs, they also introduced problems like vanishing gradients, longer training time, and higher space requirements for training and deployment. ResNet \cite{he2016deep} proposed skip-connections based on identity mapping, which reduced the optimization issues of deep networks. This allowed for using deeper and more complex networks. WideResNet \cite{zagoruyko2016wide} restricted the network depth and used wider layers to improve the performance, thereby modifying this idea. ResNeXt \cite{xie2017aggregated} proposed parallel aggregated transformations blocks and showed that increasing the number of such parallel blocks led to better performance. 

Our proposed approach improves the performance of CNNs by introducing mini-batches of data, that is different from the input data distribution, into the training routine. 
Further, it can be used along with any performance improvement techniques.

\section{Proposed Method}
Our proposed approach aims to improve network performance on a dataset by injecting passive mini-batches of data from a different data distribution, which we refer to as the passive dataset. Since our focus is on performing better on the original data, we propose a training process wherein the training is done on very few passive mini-batches as compared to the number of mini-batches of the original input data. The network is split into a base sub-network and two dataset specific sub-networks for the original and passive datasets respectively.

Suppose, the original input dataset $A=$ $\{x_{Au},y_{Au}\}_{u=1}^{N_A}$, where $N_A$ denotes the number of samples in the original dataset and the passive dataset $P=$ $\{x_{Pv},y_{Pv}\}_{v=1}^{N_P}$, where $N_P$ denotes the number of samples in the passive dataset. The base sub-network is denoted as $W$. $M_A$ and $M_P$  are used to denote the dataset specific sub-network for the original and passive datasets respectively.
$W$ has parameters $\theta_{W}$, $M_A$ has parameters $\theta_{M_A}$ and $M_P$ has parameters $\theta_{M_P}$. The full network for the original data, which will also be the final network, can be defined as 
\begin{equation}
 \hat{y}_{Au} = M_A(W(x_{Au};\theta_W);\theta_{M_A}) 
\end{equation}

where, $x_{Au}$ is the $u^{th}$ input from the original data and $\hat{y}_{Au}$ is the predicted output for that input. The full model for the passive dataset is the same, but it uses the passive dataset sub-network $M_P$ instead of $M_A$. The full model for the passive dataset can be defined as 
\begin{equation}
 \hat{y}_{Pv} = M_P(W(x_{Pv};\theta_W);\theta_{M_P}) 
\end{equation}

where $x_{Pv}$ is the $v^{th}$ input from the passive dataset, and $\hat{y}_{Pv}$ is the predicted output for that input.

The joint model has the parameters $\theta = \{\theta_W, \theta_{M_A}, \theta_{M_P} \}$. Our goal is to find the optimal $\theta^{*}$ that maximizes the performance on the original dataset such that,
\begin{equation}
 \theta^{*} = argmin_{\theta} \{ \frac{\Sigma_{u=1}^{N_A} L(y_{Au},\hat{y}_{Au})}{N_A}  + \frac{\Sigma_{v=1}^{N_P} L(y_{Pv},\hat{y}_{Pv})}{N_P}  \} 
\end{equation}
Where $L$ denotes a loss function that finds the prediction losses for the datasets. Generally, the deep learning models are optimized using some variant of stochastic gradient descent, where the training data is split into mini-batches, and the network parameters are updated using gradients based on the loss obtained for each mini-batch. Our procedure involves training with mini-batches. Let $m^{i}_A$ denote the $i^{th}$ mini-batch of $N_{mA}$ pairs $(x_{Ai},y_{Ai})$ for the original dataset, and $m^{j}_P$ denote the $j^{th}$ mini-batch of $N_{mP}$ pairs $(x_{Pj},y_{Pj})$  for the passive dataset. 

As mentioned earlier, our only goal is to maximize the model performance on the original dataset. Therefore, every time the model has been trained on $g>>1$ mini-batches of the original dataset, we train the model on one passive dataset mini-batch i.e. 

\begin{equation}
\resizebox{0.9\hsize}{!}{$\theta^{t}=\theta^{t-1} -\alpha \nabla \{ \frac{ \Sigma_{i=1}^{N_{mA}} L(y_{Ai},\hat{y}_{Ai})}{N_{mA}} +  \mathbf{I}_{\{i\%g==0\}}  \frac{\Sigma_{j=1}^{N_{mP}} L(y_{Pj},\hat{y}_{Pj})}{N_{mP}}\}$}
\end{equation}

$\theta^{t}$ is the value of the parameters after the model has trained for time $t$, $\alpha$ is the learning rate, and the indicator function $\mathbf{I}_{\{i\%g==0\}}$ denotes that the model will be trained on $1$ passive dataset mini-batch every time it trains on $g$ original dataset mini-batches (refer to Algorithm \ref{alg:algorithm}). 

\begin{algorithm}[t]
\caption{Mini-batch stochastic gradient descent training of network using the PBITT approach}
\label{alg:algorithm}
\begin{algorithmic}[1]
\FOR{number of training iterations}
\FOR{number of batches of the original dataset}
\STATE Sample a mini-batch of $N_{mA}$ pairs $(x_{Ai},y_{Ai})$ from the original dataset
\STATE Update model parameters $\theta$ along its stochastic gradient for the original dataset mini-batch (\textit{Keeping $\theta_{M_P}$ fixed}).
$$ \theta^{t} = \theta^{t-1}\ -\ \alpha \nabla_{\theta} \{ \frac{1}{N_{mA}}\Sigma_{i=1}^{N_{mA}} L(y_{Ai},\hat{y}_{Ai}) \} $$
\IF{OriginalDataMiniBatchNo i is divisible by $g$}
\STATE Sample a mini-batch of $N_{mP}$ pairs $(x_{Pj},y_{Pj})$ from the passive dataset
\STATE Update model parameters $\theta$ along its stochastic gradient for the passive dataset mini-batch (\textit{Keeping $\theta_{M_A}$ fixed}).
$$ \theta^{t} = \theta^{t-1}\ -\ \alpha \nabla_{\theta} \{ \frac{1}{N_{mP}}\Sigma_{j=1}^{N_{mP}} L(y_{Pj},\hat{y}_{Pj}) \} $$
\ENDIF

\ENDFOR
\ENDFOR
\end{algorithmic}
 \end{algorithm}

\subsection{Intuition behind PBITT}
When we train a model by using a passive dataset in addition to the original dataset, then the model has to search for a representation that captures the information needed for both of the datasets, and therefore there is very less chance of overfitting on the original dataset. 

However, since the passive dataset has a different distribution than the original dataset, the representation that the network will learn might not be discriminative enough as compared to the representations learned in the older training technique. This will result in harming the performance of the network. This is why we control the number of passive mini-batches. By keeping the proportion of mini-batches very low, we are still able to reduce the overfitting of the network on the original dataset but without negatively affecting the network performance. As a result, the model should perform better on the original dataset even though its performance on the passive dataset might not be optimal. Our approach is able to reduce the overfitting of the network, which results in better performance, and this has been experimentally shown in the ablation studies (Section \ref{subsec:overfitting}).

  \subsection{PBITT is not Multi-Task Learning}
 Our proposed training technique PBITT should not be confused with Multi-Task Learning (MTL). 
 
 Multi-Task Learning involves using the same network to learn multiple tasks on the same dataset. This allows the network to learn better features from the input and as a result, learns to perform better for all the tasks. By sharing representations between related tasks, the aim is to make the network to generalize better on all the tasks. 
 
 However, our proposed algorithm is a training technique that makes use of mini-batches from a different data distribution, which we refer to as the passive dataset, to reduce the overfitting of the network and, as a result, increase the generalization performance of the network. We are not bothered about whether the network performs well on the passive dataset. Our only focus is to make the network perform better for the original data. The performance on the passive mini-batches will, anyways, be not good since they are an extreme minority when compared to the original mini-batches.

 \subsection{PBITT is not Data Augmentation}
  Our proposed training technique PBITT should also not be confused with data augmentation. 
 
 In data augmentation, existing data from the input source is manipulated using geometric operations to create more instances of input from the same distribution. In some cases, generative models are also used to generate data from the same data distribution. This is done to fulfill the high data requirement for training deep networks.
 
 However, in our proposed technique, we use mini-batches from a different data distribution as part of the training. Since these data points are from a different data distribution, they do not increase/augment the original input data. As will be shown later, the amount of such data used is so less as compared to the original data, they also do not affect the batch statistics of the batch norm layers.

\section{Implementation}
We draw the passive mini-batches from a dataset where the images do not match the general content or pattern of the original images at a specific size. This ensures that the distribution of passive mini-batches is different from the original input data.

All the convolutional layers ($W$) are shared by both the original and passive datasets. A set of separate fully connected layers ($M_P$) is added for the passive dataset, which takes input from the last convolutional layer of the network.

After training the model on the original dataset for every $g$ mini-batches, we train the model on $1$ mini-batch of the passive dataset. 

If we use $g=1$, then we will be training on an equal number of original and passive dataset mini-batches, and this will double the training time, which is undesirable. Further, this $g=1$ may affect the performance on the original dataset \cite{li2018learning,mallya2018packnet,mallya2018piggyback,he2018multi} since data distributions are different. 

We use $g=100$ for most of our experiments, and this choice has been experimentally validated in the ablation studies section. Therefore, for every $100$ mini-batches of the original dataset, we train on $1$ passive dataset mini-batch. We also experimentally validate in our ablation studies that our approach of Passive Batch Injection Training Technique (PBITT) does not depend on the selection of the passive dataset.

During training, the dataset-specific layers for only the corresponding dataset are enabled for a given mini-batch, depending on which type of mini-batch the model is being trained upon. For example, when performing training on the original dataset, we only activate the original dataset-specific layers. The shared sub-network is always activated. After the training is done, we can simply remove the dataset-specific fully connected layers for the passive dataset and use the rest of the model for the original dataset.

\subsection{Additional Parameters}
Our implementation ensures that only a few additional parameters are added to the original model, that too, only during training. More specifically, the only additional parameters introduced are the weights and biases of the set of fully connected dataset-specific layers for the passive dataset. There are \textit{no additional} parameters in the final model since we remove the set of dataset-specific fully connected layers for the passive dataset after training.

\subsection{Additional Training Time}
Since we use a high value of $g$, the additional training time needed to train on the passive dataset is very low. When $g=100$, we use $1$ passive dataset mini-batch for every $100$ original dataset mini-batches. Therefore, the overall increase in time per original dataset epoch is about $1$ percent, which is \textit{negligible}. There is \textit{no additional} time needed for testing since we remove the set of dataset-specific fully connected layers for the passive dataset after training.

\begin{table*}[t]
    \begin{center}
    \caption{Single-crop accuracy ($\%$) on the ImageNet validation set for ResNet-50 and WideResNet-18 (widen=2).}
        \label{tab:imagenet}
        \scalebox{1.0}{
        \renewcommand{\arraystretch}{1.3}
        \addtolength{\tabcolsep}{12pt}
         \begin{tabular}{| c | c | c | c | c |}
            \hline
          \textbf{Models} &\textbf{Active} &  \textbf{Passive } & \multicolumn{2}{|c|}{\textbf{Accuracy}}\\ 
          \cline{4-5}
          \multicolumn{1}{|c|}{} & \textbf{Dataset}& \textbf{Dataset} & \textbf{Top-1} & \textbf{Top-5}  \\
          \hline\hline
          \multicolumn{1}{|l|}{ResNet-50 (Baseline)} &ImageNet & - & 76.1 &  92.9 \\
          \hline
          \multicolumn{1}{|l|}{SRIP \cite{bansal2018can}} &ImageNet& - & 76.4 &  93.1 \\
          \hline
          \multicolumn{1}{|l|}{\textbf{PBITT $g=100$ (Ours)}} &ImageNet& CIFAR-100 & \textbf{76.8} &  \textbf{93.3} \\
          \hline\hline
                    \multicolumn{1}{|l|}{WResNet-18-2 (Baseline)} &ImageNet& - & 74.4 &  91.8 \\
          \hline
          \multicolumn{1}{|l|}{SRIP \cite{bansal2018can}} &ImageNet& - & 74.6 &  92.0 \\
          \hline
          \multicolumn{1}{|l|}{\textbf{PBITT $g=100$ (Ours)}}&ImageNet & CIFAR-100 & \textbf{74.8} &  \textbf{92.2} \\
          \hline
         \end{tabular}}
     \end{center}
     
\end{table*}

\section{Experiments}
\label{experimental}
This section explores the experimental results of training networks with our proposed approach (PBITT: \textbf{P}assive \textbf{B}atch \textbf{I}njection \textbf{T}raining \textbf{T}echnique) on various datasets. We perform experiments on image classification and object detection. For the image classification, we use VGG, ResNet, and WideResNet networks over the ImageNet and CIFAR datasets. We use the Faster R-CNN network for object detection over the MS-COCO dataset.

\begin{table*}[t]

    \begin{center}
    \caption{Classification accuracy ($\%$) of ResNet-56, WideResNet-22 (widen=10), and VGG-16 on CIFAR-10 and CIFAR-100 datasets.}
        \label{tab:cifar}
         
        \addtolength{\tabcolsep}{12pt}
        \renewcommand{\arraystretch}{1.3}
         \begin{tabular}{| c | c | c | c |}
        \hline
          \textbf{Models} &  \textbf{Active } &  \textbf{Passive } & \multicolumn{1}{|c|}{\textbf{Accuracy}}\\ 
          \cline{4-4}
          \multicolumn{1}{|c|}{} & \textbf{Dataset} & \textbf{Dataset} & \textbf{Top-1}   \\
           \hline\hline
          \multicolumn{1}{|l|}{ResNet-56 (Baseline)} & CIFAR-10 & -  &  93.4 \\
          \hline
          \multicolumn{1}{|l|}{PBITT $g=1$ } & CIFAR-10 & SVHN  &  93.3 \\
          \hline
          \multicolumn{1}{|l|}{SRIP \cite{bansal2018can}} & CIFAR-10 & -  &  93.7 \\
          \hline
          \multicolumn{1}{|l|}{\textbf{PBITT $g=100$ (Ours)}} & CIFAR-10 & SVHN  &  \textbf{94.4} \\
          \hline\hline
        \multicolumn{1}{|l|}{WideResNet-22-10 (Baseline)} & CIFAR-10 & -  &  95.6 \\
          \hline
          \multicolumn{1}{|l|}{PBITT $g=1$ } & CIFAR-10 & SVHN &  95.4 \\
          \hline
          \multicolumn{1}{|l|}{SRIP \cite{bansal2018can}} & CIFAR-10 & -  &  95.8 \\
          \hline
          \multicolumn{1}{|l|}{\textbf{PBITT $g=100$ (Ours)}} & CIFAR-10 & SVHN  &  \textbf{96.0} \\
          \hline\hline
         \multicolumn{1}{|l|}{VGG-16 (Baseline)} & CIFAR-10 & -  &  93.5  \\
          \hline
          \multicolumn{1}{|l|}{PBITT $g=1$ } & CIFAR-10 & SVHN &  93.4 \\
          \hline
          \multicolumn{1}{|l|}{SRIP \cite{bansal2018can}} & CIFAR-10 & - &   93.9 \\
          \hline
          \multicolumn{1}{|l|}{\textbf{PBITT $g=100$ (Ours)}} & CIFAR-10 & SVHN  &  \textbf{94.1} \\
          \hline\hline
        \multicolumn{1}{|l|}{ResNet-56 (Baseline)} & CIFAR-100 & -  &  71.6 \\
          \hline
          \multicolumn{1}{|l|}{PBITT $g=1$ } & CIFAR-100& SVHN &   71.3 \\
          \hline
          \multicolumn{1}{|l|}{SRIP \cite{bansal2018can}} & CIFAR-100& - &   71.6 \\
          \hline
          \multicolumn{1}{|l|}{\textbf{PBITT $g=100$ (Ours)}} & CIFAR-100& SVHN  &  \textbf{71.7} \\
          \hline\hline
        \multicolumn{1}{|l|}{WideResNet-22-10 (Baseline)} & CIFAR-100& -  &  79.3  \\
          \hline
          \multicolumn{1}{|l|}{PBITT $g=1$ } & CIFAR-100& SVHN &  79.2 \\
          \hline
          \multicolumn{1}{|l|}{SRIP \cite{bansal2018can}} & CIFAR-100& - &  79.9 \\
          \hline
          \multicolumn{1}{|l|}{\textbf{PBITT $g=100$ (Ours)}} & CIFAR-100& SVHN  &  \textbf{80.2} \\
          \hline\hline
        \multicolumn{1}{|l|}{VGG-16 (Baseline)} & CIFAR-100& -  &  72.0 \\
          \hline
          \multicolumn{1}{|l|}{PBITT $g=1$ } & CIFAR-100& SVHN &   71.9 \\
          \hline
          \multicolumn{1}{|l|}{SRIP \cite{bansal2018can}} & CIFAR-100&  - &  73.8\\
          \hline
          \multicolumn{1}{|l|}{\textbf{PBITT $g=100$ (Ours)}} & CIFAR-100& SVHN  &  \textbf{74.1} \\
          \hline

         \end{tabular}
     \end{center}
     
\end{table*}

\subsection{Image Classification}
\label{classification}

We conduct experiments on ImageNet \cite{russakovsky2015imagenet}, CIFAR-10, and CIFAR-100  \cite{krizhevsky2009learning} datasets. For the ImageNet dataset, we experiments on ResNet-50 \cite{he2016deep} and WideResNet-18 (widen=2) \cite{zagoruyko2016wide}. We use the CIFAR-100 dataset as the passive dataset (for ImageNet classification) after resizing the images to $224\times224$.  For each network, the baseline is the original model. We compare our approach to a recent performance boosting method, Spectral Restricted Isometry Property (SRIP) regularization \cite{bansal2018can}, which improves the performance of state-of-the-art models by using orthogonality regularization. 

The training set of ImageNet (ILSVRC-2012) contains 1.2 million images and 1000 object categories. The validation set contains 50,000 images. For ImageNet experiments, we perform standard data augmentation methods of random cropping to a size of $224\times224$ and random horizontal flipping. For optimization, stochastic gradient descent (SGD) is used with momentum $0.9$ and a mini-batch size of $256$. Initially, the learning rate is set to $0.1$ and is decreased by a factor of $10$ for every $30$ epoch. The models are trained from scratch for $120$ epochs. For evaluation, the validation images are subjected to center cropping of size $224\times224$.

We use $g=100$, i.e. 1 passive dataset mini-batch for every 100 original dataset mini-batches. At the end of the training, the passive dataset-specific layers are removed. Therefore, our approach is able to improve the performance of deep CNNs on the original dataset without introducing any extra parameters or FLOPS  (floating-point operations per second) in the final model.
 
 Table~\ref{tab:imagenet} shows the classification accuracy on ImageNet dataset using ResNet-50 and WideResNet-18 (widen=2) networks for the original network, our proposed method PBITT ($g=100$) and a recent performance boosting method SRIP. Our method shows consistent performance improvement as compare to SRIP. Our method ($g=100$) uses CIFAR-100 as the passive dataset, i.e., for every $100$ mini-batches of ImageNet, 1 CIFAR-100 mini-batch was used for training the network. 
 
 As shown in Table~\ref{tab:imagenet}, our approach yields a significant relative performance improvement (of 0.7 \% in Top-1 accuracy) over the baseline for ResNet-50 while there is no increment in inference (test) time since we remove the set of dataset-specific fully connected layers for the passive dataset after the training is completed.

For the CIFAR-10 and CIFAR-100 datasets, we experiments on ResNet-56 \cite{he2016deep}, VGG-16 \cite{simonyan2014very}, and WideResNet-22 (widen=10) \cite{zagoruyko2016wide}. We use SVHN  \cite{netzer2011svhn} dataset as the passive dataset. For each network, the baseline is the original model. We compare our approach to a recent performance boosting method, Spectral Restricted Isometry Property (SRIP) regularization \cite{bansal2018can}.

The CIFAR-10 dataset consists of 60000 $32\times32$ color images in 10 classes, with 6000 images per class. The CIFAR-100 dataset consists of 60000 $32\times32$ color images in 100 classes, with 600 images per class. Out of these, 50000 are training images, and 10000 are test images. For CIFAR experiments, we perform standard data augmentation methods of random cropping to a size of $32\times32$ (zero-padded on each side with four pixels before taking a random crop) and random horizontal flipping. For optimization, stochastic gradient descent (SGD) is used with momentum $0.9$ and a minibatch size of $128$. Initially, the learning rate is set to $0.1$ and is decreased by a factor of $5$ for every $50$ epoch. The models are trained from scratch for $250$ epochs. For evaluation, the test images are used.

We can see from Table~\ref{tab:cifar} that PBITT $g=1$ performs even worse than the baseline since when we use $g=1$, we are training the same model equally for both the datasets. Since the network is almost fully shared and the two datasets have different data distributions, the overall performance on both has been experimentally found to be lower than their separate models by previous works such as \cite{li2018learning,mallya2018packnet,mallya2018piggyback,he2018multi}. Further results on the mini-batch proportion of $g$ have been provided in the ablation studies section.

Table~\ref{tab:cifar} shows the classification accuracy on CIFAR-10 and CIFAR-100 datasets using ResNet-56, WideResNet-22 (widen=10) and VGG-16 networks for the original networks, PBITT $(g=100)$ and SRIP. Our method PBITT $(g=100)$ shows consistent performance improvement in all cases. PBITT $(g=100)$ uses SVHN as the passive dataset i.e. for every 100 mini-batches of CIFAR-10/CIFAR-100, 1 SVHN mini-batch is used for training the network.

\subsection{Object Detection}

We use Faster R-CNN network \cite{ren2015faster} for object detection. The ResNet-50 model is used as a base model in the Faster R-CNN network \cite{ren2015faster}. We use MS-COCO dataset \cite{lin2014microsoft} for object detection. Table \ref{tab:detection} shows the performance of the object detector using the base ResNet-50 model and the modified ResNet-50 with the PBITT approach where CIFAR-100 is used as the passive dataset. The PBITT modified ResNet-50 shows improvement over the base model. We can conclude from these experiments that the PBITT approach induces substantial performance improvements in the network performance across several architectures, datasets, and tasks than other existing methods.

\begin{table}[t]
  \begin{center}
    \caption{Performance on Object Detection: Object detection mAP ($\%$) on the MS COCO dataset using Faster R-CNN network.}
    \label{tab:detection}
    \scalebox{1.0}{
    \renewcommand{\arraystretch}{1.2}
        \addtolength{\tabcolsep}{2pt}
     \begin{tabular}{|l|c|c|} 
      \hline
     \textbf{Base Model} & \textbf{AP@IoU=0.5} & \textbf{AP@IoU=0.5:0.95}\\ 
      \hline\hline
      ResNet-50 (Baseline) &  $45.2$ & $25.1$ \\
     \hline
      \textbf{ResNet-50 (PBITT)} &  $\textbf{46.0}$ & $\textbf{25.7}$ \\
     \hline
    \end{tabular}}
  \end{center}
  
\end{table}

\begin{table*}[t]
    \begin{center}
         \caption{Gap in train and test classification accuracy ($\%$) for ResNet-56 with different proportion of the training data using the baseline and proposed approach PBITT ($g=100$).}
        \label{tab:cifar10ablationoverfit}
        \scalebox{1.0}{
        \renewcommand{\arraystretch}{1.2}
        \addtolength{\tabcolsep}{5pt}
         \begin{tabular}{| c | c | c | c | c |}
            \hline
           \multicolumn{2}{|c|}{Datasets} &  \textbf{Model} & \textbf{\% Training data} & \textbf{Train-Test Acc. Gap}\\ 
          \cline{1-2}
          \cline{5-5}
           \textbf{Active} &  \textbf{Passive} & \textbf{ } & \textbf{ } & \textbf{Top-1}   \\
            \hline\hline
           CIFAR10 & - & ResNet-56 (Baseline) & 100 & 6.5 \\
          \hline
           CIFAR10 & SVHN & \textbf{ResNet-56 (PBITT)}   & 100 &  \textbf{5.5} \\
          \hline
          CIFAR10 & - & ResNet-56 (Baseline) & 25 &  16.2 \\
          \hline
           CIFAR10 &  SVHN & \textbf{ResNet-56 (PBITT)} & 25 & \textbf{14.9} \\
          \hline
           CIFAR10 & -  & ResNet-56 (Baseline) & 12.5  & 24.4 \\
          \hline
           CIFAR10 & SVHN  & \textbf{ResNet-56 (PBITT)} & 12.5 & \textbf{21.1} \\
          \hline
         \end{tabular}}
     \end{center}
     
\end{table*}

\section{Ablation Studies}
We validate the proposed approach using extensive ablation studies. We perform ablation experiments to check the effects of our approach on the model overfitting, to check the effect of the choice of $g$ for PBITT, the effect of swapping of the original and passive dataset, and the effect of changing the passive dataset.

\subsection{Effect of PBITT on Overfitting}\label{subsec:overfitting}
We perform experiments to show how much the model overfits under the original training procedure and our proposed PBITT approach. We use the difference between the training and test accuracy as a measure of overfitting. Higher the difference/gap, higher is the overfitting on the training data. We perform training with 100\%, 25\%, and 12.5\% of the training data. From Table \ref{tab:cifar10ablationoverfit}, we can see that the gap between training and test classification accuracy on the CIFAR-10 dataset is consistently lower in PBITT ($g=100$) as compared to the baseline. 

We should note here that the ResNet-56 model already uses \textit{batch normalization}, and the standard training procedure uses \textit{weight decay ($L_2$ regularization)}, both of which are used to reduce overfitting. Therefore, from the results, we can see that our approach further reduces the level of overfitting on top of existing techniques to reduce overfitting. This results in better generalization and hence, better test accuracy. This is the reason why the models trained using our approach performs better than the original models on the same dataset.

\begin{table*}[t]
  \begin{center}
    \caption{Classification accuracy (\%) for ResNet-56 with SVHN and GTSRB as the passive dataset ($g=100$). Our approach of Passive Batch Injection Training Technique (BITT  $g=100$) does not depend on the selection of passive dataset for performance improvement.}
    \label{tab:ablationmultidataset}
    \scalebox{1.0}{
    \renewcommand{\arraystretch}{1.2}
        \addtolength{\tabcolsep}{12pt}
     \begin{tabular}{|c|c|c|c|} 
      \hline
      \multicolumn{2}{|c|}{Datasets} &  \textbf{Models} & \textbf{Active Acc.}\\ 
     \cline{1-2}
     \cline{4-4}
    \textbf{Active} &  \textbf{Passive} & \textbf{ } & \textbf{Top-1}   \\
      \hline\hline
    CIFAR-10 & - &  ResNet-56 (Baseline) &  $93.4$  \\
     \hline
      CIFAR-10 & \textbf{SVHN} &  \textbf{ResNet-56 (PBITT)} &  $\textbf{94.4}$  \\
     \hline
      CIFAR-10 & \textbf{GTSRB} &  \textbf{ResNet-56 (PBITT)} &  $\textbf{94.3}$  \\
      \hline
    \end{tabular}}
  \end{center}
  
\end{table*}

\begin{table*}[t]
  \begin{center}
    \caption{Classification accuracy (\%) for ResNet-50 with CIFAR-100 and SVHN as the passive dataset ($g=100$).}
    \label{tab:ablationmultidataset2}
    \scalebox{1.0}{
    \renewcommand{\arraystretch}{1.2}
        \addtolength{\tabcolsep}{12pt}
     \begin{tabular}{|c|c|c|c|} 
      \hline
      \multicolumn{2}{|c|}{Datasets} &  \textbf{Models} & \textbf{Active Acc.}\\ 
     \cline{1-2}
     \cline{4-4}
    \textbf{Active} &  \textbf{Passive} & \textbf{ } & \textbf{Top-1}   \\
      \hline\hline
    ImageNet & - &  ResNet-50 (Baseline) &  $76.1$  \\
     \hline
      ImageNet & \textbf{CIFAR-100} &  \textbf{ResNet-50 (PBITT)} &  $\textbf{76.8}$  \\
     \hline
      ImageNet & \textbf{SVHN} &  \textbf{ResNet-50 (PBITT)} &  $\textbf{76.7}$  \\
      \hline
    \end{tabular}}
  \end{center}
  
\end{table*}

\begin{table}[t]
    \begin{center}
         \caption{Classification accuracy ($\%$) for ResNet-56 with different mini-batch proportion $g$.}
        \label{tab:cifar10ablationg}
        \renewcommand{\arraystretch}{1.2}
        \addtolength{\tabcolsep}{2pt}
         \begin{tabular}{| c | c | c | c |}
            \hline
           \multicolumn{2}{|c|}{Datasets} &  \textbf{g:1} & \textbf{Active Acc.}\\ 
          \cline{1-2}
          \cline{4-4}
           \textbf{Active} &  \textbf{Passive} & \textbf{ } & \textbf{Top-1}   \\
            \hline\hline
           CIFAR10 & - & 1:0 & 93.4 \\
          \hline
           CIFAR10 & SVHN & 1:1  &  93.3 \\
          \hline
          CIFAR10 & SVHN & 10:1 &  94.1 \\
          \hline
           CIFAR10 &  SVHN & \textbf{100:1} & \textbf{94.4} \\
          \hline
           CIFAR10 & SVHN  &  1000:1 & 93.6 \\
          \hline\hline
           SVHN & - & 1:0 & 96.3 \\
           \hline
           SVHN & CIFAR10 & 1:1  &  96.1 \\
           \hline
          SVHN & CIFAR10 & 10:1 &  96.7 \\
          \hline
           SVHN &  CIFAR10 & \textbf{100:1} & \textbf{96.8} \\
          \hline
           SVHN & CIFAR10  &  1000:1 & 96.4 \\
          \hline

         \end{tabular}
     \end{center}
     
\end{table}

\subsection{Selection of Passive Dataset}

We also analyze the effect of changing the passive dataset in our ablation studies to show that our approach (BITT  $g=100$) does not depend on the selection of a passive dataset.

From Table~\ref{tab:ablationmultidataset}, we can see that even when we use the German Traffic Sign Recognition Benchmark (GTRSB) dataset \cite{Stallkamp-IJCNN-2011} as the passive dataset, the PBITT ($g=100$) approach still gets better classification accuracy on the CIFAR-10 dataset. So for both SVHN and GTRSB passive datasets, which have significantly different data-distribution than CIFAR-10, our PBITT approach is able to improve the performance of the model on CIFAR-10. A similar pattern has been observed for ResNet-50 on ImageNet as shown in Table~\ref{tab:ablationmultidataset2}. The proposed approach is, therefore, a widely usable and efficient way of improving the performance of Deep CNN models.

\subsection{Mini-batch Proportion}

Our framework uses $g=100$, i.e., for every $100$ mini-batches of the original dataset 1 mini-batch of the passive dataset is used to train the model. We experimented with $g=1,10,100,1000$. We performed the experiments on the ResNet-56 network with the original dataset as a classification on the CIFAR-10 dataset and the passive dataset as a classification on the SVHN dataset.

 From Table \ref{tab:cifar10ablationg}, we can see the best original dataset performance on CIFAR-10 is obtain when we use $g=100$ i.e. 1 SVHN (passive) mini-batch for every 100 CIFAR-10 (original) mini-batch. We also note that $g=1000$ performs worse than $g=10,100$, which shows that when the proportion of the passive dataset mini-batches is extremely small, the performance on the original dataset is similar to the original network with no passive dataset $(1:0)$. This validates our choice of $g=100$.

From Table \ref{tab:cifar10ablationg}, we can see that PBITT $g=1$ performs even worse than the baseline. This is because, when we use $g=1$, we are giving equal training for both the datasets. Since the network is almost fully shared and we are using different data distributions for training, the overall performance on both datasets has been experimentally found to be lower than their separate models by previous works such as \cite{li2018learning,mallya2018packnet,mallya2018piggyback,he2018multi}. Further, $g=1$ is also not preferable because it doubles the overall training time since we have 1 passive dataset mini-batch for every original dataset mini-batch.

\subsection{Swapping Active and Passive Datasets}

From Table~\ref{tab:cifar10ablationg}, we can see that even when we swap the original and passive dataset, the new original dataset (SVHN) behaves similarly to the previous original dataset (CIFAR-10), i.e., the network shows the best original dataset performance when it uses $g=100$ and the worst performance for $g=1$. This further validates the usability of our approach.

\section{Conclusion}
In this paper, we have proposed a Passive Batch Injection Training Technique (PBITT) approach to improve the performance of networks on an original dataset by additionally training on a passive dataset mini-batch after every substantial number of original dataset mini-batches. Our approach reduces the level of overfitting further, even on top of $L_2$ regularization and batch normalization, which are standard techniques to reduce overfitting. Our approach does not increase the number of parameters and inference (test) time of the final model while improving the performance of deep CNNs. We thoroughly evaluate the proposed approach on several standard architectures: VGG, ResNet, and WideResNet, and on several popular datasets: CIFAR-10, CIFAR-100, SVHN, and ImageNet. On using the proposed approach, we observe consistent accuracy improvements. We also show that our approach generalizes to detection as well. We validate our proposed approach through various ablation studies (effect of PBITT on overfitting,  mini-batch proportion, swapping active and passive datasets, and selection of the passive dataset). Therefore, we can conclude that Passive Batch Injection Training Technique is an efficient way of improving the performance of Deep CNN models.

{\small
\bibliographystyle{IEEEtran}
\bibliography{egbib}

\begin{thebibliography}{10}
\providecommand{\url}[1]{#1}
\csname url@samestyle\endcsname
\providecommand{\newblock}{\relax}
\providecommand{\bibinfo}[2]{#2}
\providecommand{\BIBentrySTDinterwordspacing}{\spaceskip=0pt\relax}
\providecommand{\BIBentryALTinterwordstretchfactor}{4}
\providecommand{\BIBentryALTinterwordspacing}{\spaceskip=\fontdimen2\font plus
\BIBentryALTinterwordstretchfactor\fontdimen3\font minus
  \fontdimen4\font\relax}
\providecommand{\BIBforeignlanguage}[2]{{%
\expandafter\ifx\csname l@#1\endcsname\relax
\typeout{** WARNING: IEEEtran.bst: No hyphenation pattern has been}%
\typeout{** loaded for the language `#1'. Using the pattern for}%
\typeout{** the default language instead.}%
\else
\language=\csname l@#1\endcsname
\fi
#2}}
\providecommand{\BIBdecl}{\relax}
\BIBdecl

\bibitem{glorot2010understanding}
X.~Glorot and Y.~Bengio, ``Understanding the difficulty of training deep
  feedforward neural networks,'' in \emph{Proceedings of the thirteenth
  international conference on artificial intelligence and statistics}, 2010,
  pp. 249--256.

\bibitem{ioffe2015batch}
S.~Ioffe and C.~Szegedy, ``Batch normalization: Accelerating deep network
  training by reducing internal covariate shift,'' in \emph{International
  Conference on Machine Learning}, 2015, pp. 448--456.

\bibitem{dauphin2014identifying}
Y.~N. Dauphin, R.~Pascanu, C.~Gulcehre, K.~Cho, S.~Ganguli, and Y.~Bengio,
  ``Identifying and attacking the saddle point problem in high-dimensional
  non-convex optimization,'' in \emph{Advances in neural information processing
  systems}, 2014, pp. 2933--2941.

\bibitem{tetko1995neural}
I.~V. Tetko, D.~J. Livingstone, and A.~I. Luik, ``Neural network studies. 1.
  comparison of overfitting and overtraining,'' \emph{Journal of chemical
  information and computer sciences}, vol.~35, no.~5, pp. 826--833, 1995.

\bibitem{srivastava2014dropout}
N.~Srivastava, G.~Hinton, A.~Krizhevsky, I.~Sutskever, and R.~Salakhutdinov,
  ``Dropout: a simple way to prevent neural networks from overfitting,''
  \emph{The Journal of Machine Learning Research}, vol.~15, no.~1, pp.
  1929--1958, 2014.

\bibitem{saxe2013exact}
A.~M. Saxe, J.~L. McClelland, and S.~Ganguli, ``Exact solutions to the
  nonlinear dynamics of learning in deep linear neural networks,'' \emph{arXiv
  preprint arXiv:1312.6120}, 2013.

\bibitem{he2016deep}
K.~He, X.~Zhang, S.~Ren, and J.~Sun, ``Deep residual learning for image
  recognition,'' in \emph{Proceedings of the IEEE conference on computer vision
  and pattern recognition}, 2016, pp. 770--778.

\bibitem{bansal2018can}
N.~Bansal, X.~Chen, and Z.~Wang, ``Can we gain more from orthogonality
  regularizations in training deep networks?'' in \emph{Advances in Neural
  Information Processing Systems}, 2018, pp. 4261--4271.

\bibitem{ng2004feature}
A.~Y. Ng, ``Feature selection, l 1 vs. l 2 regularization, and rotational
  invariance,'' in \emph{Proceedings of the twenty-first international
  conference on Machine learning}.\hskip 1em plus 0.5em minus 0.4em\relax ACM,
  2004, p.~78.

\bibitem{van2017l2}
T.~van Laarhoven, ``L2 regularization versus batch and weight normalization,''
  \emph{arXiv preprint arXiv:1706.05350}, 2017.

\bibitem{singh2019hetconv}
P.~Singh, V.~K. Verma, P.~Rai, and V.~P. Namboodiri, ``Hetconv: Heterogeneous
  kernel-based convolutions for deep cnns,'' in \emph{Proceedings of the IEEE
  Conference on Computer Vision and Pattern Recognition}, 2019, pp. 4835--4844.

\bibitem{singh2019stability}
P.~Singh, V.~S.~R. Kadi, N.~Verma, and V.~P. Namboodiri, ``Stability based
  filter pruning for accelerating deep cnns,'' in \emph{2019 IEEE Winter
  Conference on Applications of Computer Vision (WACV)}.\hskip 1em plus 0.5em
  minus 0.4em\relax IEEE, 2019, pp. 1166--1174.

\bibitem{singh2018leveraging}
P.~Singh, V.~K. Verma, P.~Rai, and V.~Namboodiri, ``Leveraging filter
  correlations for deep model compression,'' in \emph{The IEEE Winter
  Conference on Applications of Computer Vision}, 2020, pp. 835--844.

\bibitem{singh2019multi}
P.~Singh, R.~Manikandan, N.~Matiyali, and V.~Namboodiri, ``Multi-layer pruning
  framework for compressing single shot multibox detector,'' in \emph{2019 IEEE
  Winter Conference on Applications of Computer Vision (WACV)}.\hskip 1em plus
  0.5em minus 0.4em\relax IEEE, 2019, pp. 1318--1327.

\bibitem{singh2019accuracy}
P.~Singh, P.~Mazumder, and V.~Namboodiri, ``Accuracy booster: Performance
  boosting using feature map re-calibration,'' in \emph{The IEEE Winter
  Conference on Applications of Computer Vision}, 2020, pp. 884--893.

\bibitem{singh2019falf}
P.~Singh, V.~S.~R. Kadi, and V.~P. Namboodiri, ``Falf convnets: Fatuous
  auxiliary loss based filter-pruning for efficient deep cnns,'' \emph{Image
  and Vision Computing}, p. 103857, 2019.

\bibitem{singh2019hetconvijcv}
P.~Singh, V.~K. Verma, P.~Rai, and V.~P. Namboodiri, ``Hetconv: Beyond
  homogeneous convolution kernels for deep cnns,'' \emph{International Journal
  of Computer Vision}, pp. 1--21, 2019.

\bibitem{mazumder2019cpwc}
P.~{Mazumder}, P.~{Singh}, and V.~{Namboodiri}, ``Cpwc: Contextual point wise
  convolution for object recognition,'' in \emph{ICASSP 2020 - 2020 IEEE
  International Conference on Acoustics, Speech and Signal Processing
  (ICASSP)}, 2020, pp. 4152--4156.

\bibitem{singh2019play}
P.~Singh, V.~K. Verma, P.~Rai, and V.~P. Namboodiri, ``Play and prune: Adaptive
  filter pruning for deep model compression,'' \emph{International Joint
  Conference on Artificial Intelligence (IJCAI)}, 2019.

\bibitem{li2018learning}
Z.~Li and D.~Hoiem, ``Learning without forgetting,'' \emph{IEEE transactions on
  pattern analysis and machine intelligence}, vol.~40, no.~12, pp. 2935--2947,
  2018.

\bibitem{mallya2018packnet}
A.~Mallya and S.~Lazebnik, ``Packnet: Adding multiple tasks to a single network
  by iterative pruning,'' in \emph{Proceedings of the IEEE Conference on
  Computer Vision and Pattern Recognition}, 2018, pp. 7765--7773.

\bibitem{mallya2018piggyback}
A.~Mallya, D.~Davis, and S.~Lazebnik, ``Piggyback: Adapting a single network to
  multiple tasks by learning to mask weights,'' in \emph{Proceedings of the
  European Conference on Computer Vision (ECCV)}, 2018, pp. 67--82.

\bibitem{he2018multi}
X.~He, Z.~Zhou, and L.~Thiele, ``Multi-task zipping via layer-wise neuron
  sharing,'' in \emph{Advances in Neural Information Processing Systems}, 2018,
  pp. 6016--6026.

\bibitem{he2015delving}
K.~He, X.~Zhang, S.~Ren, and J.~Sun, ``Delving deep into rectifiers: Surpassing
  human-level performance on imagenet classification,'' in \emph{Proceedings of
  the IEEE international conference on computer vision}, 2015, pp. 1026--1034.

\bibitem{salimans2016weight}
T.~Salimans and D.~P. Kingma, ``Weight normalization: A simple
  reparameterization to accelerate training of deep neural networks,'' in
  \emph{Advances in Neural Information Processing Systems}, 2016, pp. 901--909.

\bibitem{pascanu2013difficulty}
R.~Pascanu, T.~Mikolov, and Y.~Bengio, ``On the difficulty of training
  recurrent neural networks,'' in \emph{International conference on machine
  learning}, 2013, pp. 1310--1318.

\bibitem{dorobantu2016dizzyrnn}
V.~Dorobantu, P.~A. Stromhaug, and J.~Renteria, ``Dizzyrnn: Reparameterizing
  recurrent neural networks for norm-preserving backpropagation,'' \emph{arXiv
  preprint arXiv:1612.04035}, 2016.

\bibitem{arjovsky2016unitary}
M.~Arjovsky, A.~Shah, and Y.~Bengio, ``Unitary evolution recurrent neural
  networks,'' in \emph{International Conference on Machine Learning}, 2016, pp.
  1120--1128.

\bibitem{mhammedi2017efficient}
Z.~Mhammedi, A.~Hellicar, A.~Rahman, and J.~Bailey, ``Efficient orthogonal
  parametrisation of recurrent neural networks using householder reflections,''
  in \emph{Proceedings of the 34th International Conference on Machine
  Learning-Volume 70}.\hskip 1em plus 0.5em minus 0.4em\relax JMLR. org, 2017,
  pp. 2401--2409.

\bibitem{vorontsov2017orthogonality}
E.~Vorontsov, C.~Trabelsi, S.~Kadoury, and C.~Pal, ``On orthogonality and
  learning recurrent networks with long term dependencies,'' in
  \emph{Proceedings of the 34th International Conference on Machine
  Learning-Volume 70}.\hskip 1em plus 0.5em minus 0.4em\relax JMLR. org, 2017,
  pp. 3570--3578.

\bibitem{wisdom2016full}
S.~Wisdom, T.~Powers, J.~Hershey, J.~Le~Roux, and L.~Atlas, ``Full-capacity
  unitary recurrent neural networks,'' in \emph{Advances in Neural Information
  Processing Systems}, 2016, pp. 4880--4888.

\bibitem{huang2017densely}
G.~Huang, Z.~Liu, L.~Van Der~Maaten, and K.~Q. Weinberger, ``Densely connected
  convolutional networks,'' in \emph{Proceedings of the IEEE conference on
  computer vision and pattern recognition}, 2017, pp. 4700--4708.

\bibitem{zoph2018learning}
B.~Zoph, V.~Vasudevan, J.~Shlens, and Q.~V. Le, ``Learning transferable
  architectures for scalable image recognition,'' in \emph{Proceedings of the
  IEEE conference on computer vision and pattern recognition}, 2018, pp.
  8697--8710.

\bibitem{szegedy2015going}
C.~Szegedy, W.~Liu, Y.~Jia, P.~Sermanet, S.~Reed, D.~Anguelov, D.~Erhan,
  V.~Vanhoucke, and A.~Rabinovich, ``Going deeper with convolutions,'' in
  \emph{Proceedings of the IEEE conference on computer vision and pattern
  recognition}, 2015, pp. 1--9.

\bibitem{simonyan2014very}
K.~Simonyan and A.~Zisserman, ``Very deep convolutional networks for
  large-scale image recognition,'' \emph{arXiv preprint arXiv:1409.1556}, 2014.

\bibitem{zagoruyko2016wide}
S.~Zagoruyko and N.~Komodakis, ``Wide residual networks,'' in \emph{British
  Machine Vision Conference 2016}.\hskip 1em plus 0.5em minus 0.4em\relax
  British Machine Vision Association, 2016.

\bibitem{xie2017aggregated}
S.~Xie, R.~Girshick, P.~Doll{\'a}r, Z.~Tu, and K.~He, ``Aggregated residual
  transformations for deep neural networks,'' in \emph{Computer Vision and
  Pattern Recognition (CVPR), 2017 IEEE Conference on}.\hskip 1em plus 0.5em
  minus 0.4em\relax IEEE, 2017, pp. 5987--5995.

\bibitem{russakovsky2015imagenet}
O.~Russakovsky, J.~Deng, H.~Su, J.~Krause, S.~Satheesh, S.~Ma, Z.~Huang,
  A.~Karpathy, A.~Khosla, M.~Bernstein \emph{et~al.}, ``Imagenet large scale
  visual recognition challenge,'' \emph{International journal of computer
  vision}, vol. 115, no.~3, pp. 211--252, 2015.

\bibitem{krizhevsky2009learning}
A.~Krizhevsky and G.~Hinton, ``Learning multiple layers of features from tiny
  images,'' Citeseer, Tech. Rep., 2009.

\bibitem{netzer2011svhn}
Y.~Netzer, T.~Wang, A.~Coates, A.~Bissacco, B.~Wu, and A.~Y. Ng, ``Reading
  digits in natural images with unsupervised feature learning.''\hskip 1em plus
  0.5em minus 0.4em\relax NIPSW, 2011.

\bibitem{ren2015faster}
S.~Ren, K.~He, R.~Girshick, and J.~Sun, ``Faster r-cnn: Towards real-time
  object detection with region proposal networks,'' in \emph{Advances in neural
  information processing systems}, 2015, pp. 91--99.

\bibitem{lin2014microsoft}
T.-Y. Lin, M.~Maire, S.~Belongie, J.~Hays, P.~Perona, D.~Ramanan,
  P.~Doll{\'a}r, and C.~L. Zitnick, ``Microsoft coco: Common objects in
  context,'' in \emph{European conference on computer vision}.\hskip 1em plus
  0.5em minus 0.4em\relax Springer, 2014, pp. 740--755.

\bibitem{Stallkamp-IJCNN-2011}
J.~Stallkamp, M.~Schlipsing, J.~Salmen, and C.~Igel, ``The {G}erman {T}raffic
  {S}ign {R}ecognition {B}enchmark: A multi-class classification competition,''
  in \emph{IEEE International Joint Conference on Neural Networks}, 2011, pp.
  1453--1460.

\end{thebibliography}
}

\end{document}